\documentclass[10pt,twocolumn,a4paper]{article}
\usepackage{omni_paper}
\usepackage{amsmath,graphicx,booktabs,array,colortbl}
\usepackage{balance}
\usepackage{xurl}
\usepackage[hidelinks]{hyperref}
\graphicspath{{figures/current/}{figures/}}
\definecolor{ourstint}{RGB}{235,246,242}
\definecolor{oursgreen}{RGB}{0,139,120}
\newcommand{\ours}{\textsc{Ours}}
\newcommand{\eps}{\varepsilon}
\hypersetup{pdftitle={OmniSLAM: Evidence-Authorized Revision of Persistent Object Maps},pdfauthor={Haoxiang Jie, Yaoyuan Yan, Zhiyou Heng, Liang Li, Zhaosong Li, Lifeng Liu}}
\title{OmniSLAM: Evidence-Authorized Revision\\of Persistent Object Maps}
\author{Haoxiang Jie\textsuperscript{1,*,$\dagger$}\quad
Yaoyuan Yan\textsuperscript{1,*}\quad
Zhiyou Heng\textsuperscript{1}\quad
Liang Li\textsuperscript{1}\quad
Zhaosong Li\textsuperscript{2}\quad
Lifeng Liu\textsuperscript{3}\\[5pt]
{\normalsize \textsuperscript{1}AI Lab, Country Garden Services}\\
{\normalsize \textsuperscript{2}School of Mechanical Science and Engineering, Huazhong University of Science and Technology}\\
{\normalsize \textsuperscript{3}Omni AI}}
\quad
Yaoyuan Yan\textsuperscript{*}\quad
Zhiyou Heng\quad
Liang Li\quad
Zhaosong Li\quad
Lifeng Liu}
}
\AddToHook{cmd/maketitle/after}{%
  \begingroup
  \footnotetext[1]{These authors contributed equally (co-first authors).}%
  \footnotetext[2]{Corresponding author:
    \href{mailto:jiehaoxiang@bgyfw.com}{\texttt{jiehaoxiang@bgyfw.com}}.}%
  \endgroup
}
\date{}
\begin{document}
\maketitle
\begin{abstract}
A new surface observation can extend an object map while contradicting an earlier object assignment. Treating both updates as a single merge can join unrelated objects; keeping all earlier assignments leaves fragmented objects unresolved. We present a revision module that makes these decisions separately. Historical merges require shared-mask evidence to outweigh separate-object explanations across views, including after removal of any one observation. New measured points are assigned through local competition among object seeds supported by image masks. On two preselected KITTI-360 windows, the method improves mean building/car instance AP50:90 from 17.259 to 18.762. On four further spatial holdouts, AP rises from 20.263 to 23.171, although unconditional revision reaches 23.251. Controlled comparisons attribute the gains to protecting existing identities, consolidating selected fragments, and assigning new points locally. The results also show where conservative merge decisions suppress useful updates. A garage recording demonstrates the same revision module on LiDAR--inertial--visual mapping outputs.
\par\noindent Code: {\urlstyle{same}\url{https://github.com/YanYaoyuan/omni_slam}}.
\end{abstract}

\section{Introduction}
An object mapper may reconstruct the two sides of a building under different identities. A later view can connect their measured surfaces, but geometric connectivity alone cannot determine whether their identities should be merged. The same component may include points from a neighboring structure. A revision rule must distinguish evidence for extending a surface from evidence for changing an established object assignment.

Image-based object mapping systems associate masks across views and accumulate their 3D support~\cite{cg,mc}. Their outputs provide a useful starting map, yet association errors can persist after fusion. Point filtering can remove unsupported geometry; repairing an identity assignment requires deciding which existing object should receive the affected points. This decision becomes particularly important when a broad surface proposal overlaps several persistent objects.

We propose to test historical merges against the observations that support them. For each proposed pair, we compare masks that cover both identities with masks that explain them separately. A merge is permitted only when the shared explanation remains stronger after removing any one jointly observed frame. Pairwise checks within the accepted group prevent a chain of plausible links from joining incompatible objects. A local graph then assigns new measured points and checks whether each proposed historical edit agrees with nearby object evidence. Coordinates remain tied to the original measurements.

We evaluate this revision policy within OmniSLAM, using a fixed object-mapping backbone and pose source. The experiments distinguish the effects of changing old labels, protecting them from an overwrite, and assigning newly added points. We report two prospective KITTI-360 windows separately from four spatial holdouts in previously used sequences. The main gains come from buildings; the ablations and object-level comparisons identify both useful revisions and beneficial merges that the policy rejects.

\begin{figure*}[t]
\centering
\includegraphics[width=\textwidth]{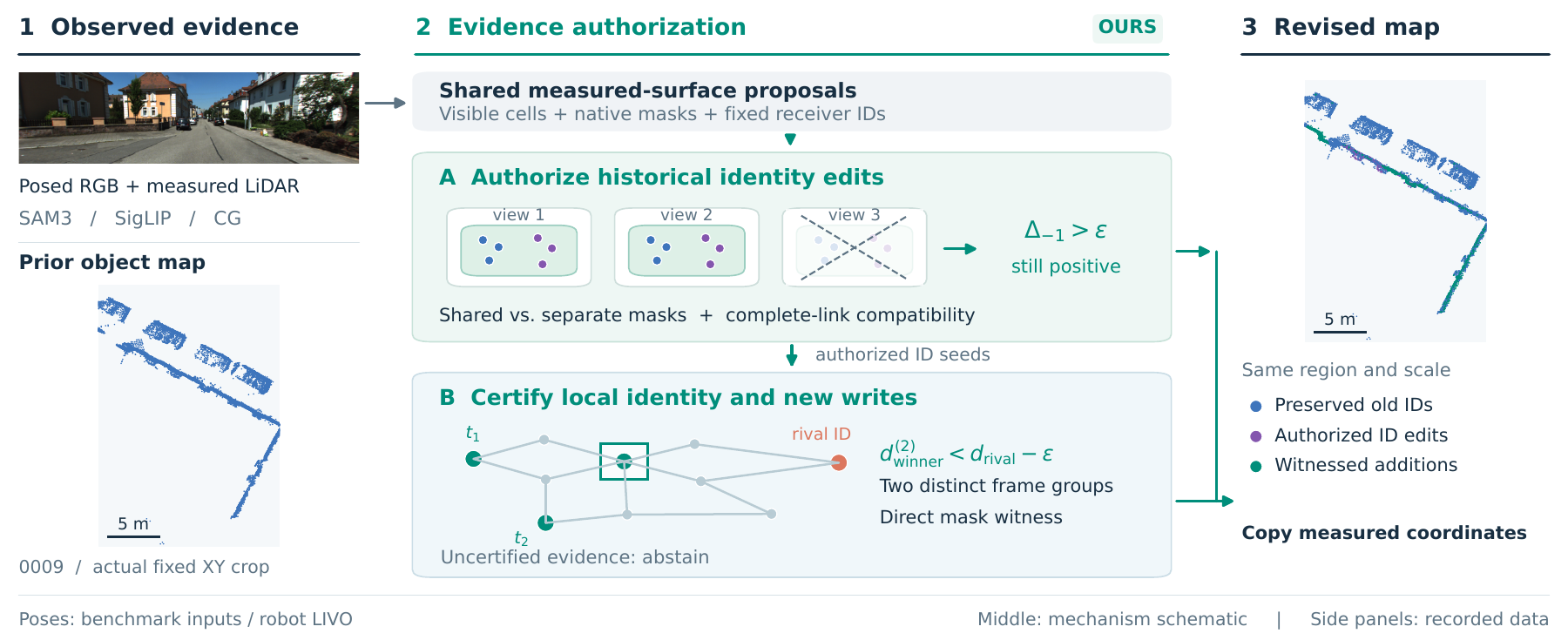}
\caption{Object-map revision. A shared proposal stage selects measured surfaces. Multi-view mask evidence determines which historical identities may merge, and local seed competition determines point assignments. Blue points retain their IDs, purple points change IDs, and teal points are added measurements. The input image and map crops are from KITTI-360 sequence 0009; the middle panels illustrate the decision rules. Map crops share their scale and bounds.}
\label{fig:architecture}
\end{figure*}

\section{Related Work}
\paragraph{Object-centric mapping.}
ConceptGraphs lifts image regions to 3D and associates them into an object graph~\cite{cg}. MaskClustering groups masks using view consensus~\cite{mc}. OVO integrates semantic mapping with SLAM~\cite{ovo}; OVI-MAP separates instance reconstruction from semantic inference~\cite{ovimap}; and VOIM accumulates soft semantic evidence before assigning instances~\cite{voim}. These approaches address how observations form a map. We consider the later revision of point-to-object assignments in a map with persistent identities.

\paragraph{Multi-view refinement.}
Point-Level Refinable Semantic Mapping updates point confidence from temporal mask, depth, and visibility evidence~\cite{refinable}. Our method uses competing mask explanations to test a proposed identity transition, followed by a local check on the affected points. The distinction matters when removing a spurious point and moving a valid point to another object require different actions.

\paragraph{Multimodal fusion.}
FusionMapper combines multimodal and temporal information for vectorized map construction~\cite{jie2025fusionmapper}, while Sparse Deep Interaction Fusion addresses 3D object detection~\cite{yan2025sparse}. We use a fixed perception stack and study the map updates supported by its accumulated observations. SAM3~\cite{sam3}, SigLIP~\cite{siglip}, and SegFormer~\cite{segformer} provide masks, descriptors, and semantic evidence. FAST-LIVO2~\cite{livo} provides poses and measured geometry for the robot experiment.

\section{Object-Map Revision}
\subsection{Map state and surface proposals}
The prior map is $\mathcal M=\{(x_i,o_i)\}$, where $x_i$ is a measured 3D point and $o_i$ is its persistent object ID. The observation archive stores source coordinates, acquisition frames, poses, and image masks. The backbone uses ConceptGraphs-style association with the frozen history2 configuration, SAM3 masks, SigLIP descriptors, and SegFormer semantic evidence. All compared revision variants receive the same prior map and cached observations.

Surface proposals are connected components of a same-class graph on 0.20\,m cells, using 16-neighbor proposals with edge lengths strictly below 0.40\,m. Visible cells contribute equal mass, divided among their projected source pixels. A shared mask-coverage score selects components supported by at least three frames; its full definition is given in Appendix~\ref{app:proposals}. Receivers are fixed by same-class overlap with prior IDs. A component without an old footprint may receive a new ID under the shared source-count and view-support rules. A component that overlaps old objects can only use the proposed existing identities.

\subsection{Historical merge decisions}
A cell represents an old identity only when all its old points have the same nonzero ID. Mixed-ID cells abstain. For identities $a$ and $b$ visible in frame $t$, let $A_{tq},B_{tq}\in[0,1]$ be the fractions of their visible footprint mass covered by same-class query mask $q$, with score $s_{tq}$. We compare a shared-mask explanation with separate explanations:
\begin{align}
S_t(a,b)&=\max_q s_{tq}\min(A_{tq},B_{tq}),\\
E_t(a\mid b)&=\max_q s_{tq}A_{tq}(1-B_{tq}),\\
R_t(a,b)&=\sqrt{E_t(a\mid b)E_t(b\mid a)}.
\end{align}
Let $\delta_t=S_t-R_t$ and $\eps=10^{-10}$. A pair is compatible when
\begin{equation}
\Delta_{-1}(a,b)=\sum_t\delta_t-\max\{0,\max_t\delta_t\}>\eps.
\label{eq:margin}
\end{equation}
This is the minimum margin after removing no frame or any one observed frame. Missing joint views contribute no evidence, and duplicate frame IDs count once.

For each fixed receiver, donor identities are considered in decreasing order of $\Delta_{-1}$, with stable-ID tie breaking. A donor joins the group only if it is compatible with every accepted member. This complete-link rule blocks transitive merging through an incompatible pair. An old point is relabeled only when its donor-to-receiver transition was proposed, the donor was accepted, and the local assignment below agrees with the receiver. Otherwise its old label is retained.

\begin{table*}[t]
\centering
\begin{minipage}[t]{.43\textwidth}\vspace{0pt}\centering\small
\textbf{(a) Prospective windows}\par\smallskip
\begin{tabular}{lrrr}
\toprule
Method & 0009 & 0010 & Mean \\
\midrule
Strong baseline & 23.098 & 11.420 & 17.259 \\
Unconditional & 22.680 & 11.420 & 17.050 \\
History-only & 22.615 & 12.809 & 17.712 \\
\textbf{Ours} & 23.327 & 14.198 & 18.762 \\
\bottomrule
\end{tabular}

\end{minipage}\hfill
\begin{minipage}[t]{.55\textwidth}\vspace{0pt}\centering\small
\textbf{(b) Spatial holdouts in previously used sequences}\par\smallskip
\begin{tabular}{lrrrrr}
\toprule
Method & 0003 & 0004 & 0005 & 0006 & Mean \\
\midrule
Strong baseline & 0.000 & 39.522 & 26.054 & 15.476 & 20.263 \\
Unconditional & 0.000 & 37.309 & 26.778 & 28.917 & 23.251 \\
History-only & 0.000 & 36.200 & 26.778 & 28.917 & 22.974 \\
Witness + history & 0.000 & 38.536 & 26.506 & 26.702 & 22.936 \\
Nearest + history & 0.000 & 38.966 & 26.913 & 26.806 & 23.171 \\
\rowcolor{ourstint}
\textbf{Ours} & 0.000 & 38.966 & 26.913 & 26.806 & 23.171 \\
MaskClustering core & 0.000 & 0.213 & 0.000 & 0.000 & 0.053 \\
\bottomrule
\end{tabular}

\end{minipage}
\caption{Instance AP50:90. The method is fixed across windows. Each mean averages the windows in its own group; missing classes are excluded from a window's class mean. MaskClustering in (b) uses a shared-source core adaptation (Appendix~\ref{app:external}). The nearest-seed control changes only new-point assignment; historical decisions remain those of Ours.}
\label{tab:results}
\end{table*}

\subsection{Local point assignment}
Within an accepted component, pure old-ID cells provide seeds when a selected mask observes them. Approved donors use the receiver's identity; other old identities remain separate. Components with no old footprint use their proposed new identity at witnessed cells. Each seed retains the frame that supports it.

Let $d_o(v)$ be the shortest-path distance from node $v$ to seeds of identity $o$. For the unique nearest identity $o^*$, let $d^{(2)}_{o^*,t}(v)$ denote its second-nearest distance from a distinct seed-frame group. The assignment passes the stability check when two such groups exist and
\begin{equation}
d^{(2)}_{o^*,t}(v)<\min_{o\ne o^*}d_o(v)-\eps.
\label{eq:certificate}
\end{equation}
Deleting any one seed-frame group then leaves a winning seed closer than every competing identity. This property assumes a fixed graph and edge weights; it does not imply that the winning label is semantically correct.

A new point is admitted when its component is accepted, its cell has direct mask support, and Eq.~\ref{eq:certificate} supplies an identity. Support is checked at cell level, so another measurement in the same 0.20\,m cell may be admitted. Coordinates are copied from the measurement archive. The shared proposal stage excludes additions within 0.05\,m of prior geometry. All variants inherit the same class scores and receive no additional variant-specific filtering. The effect of the stability check is separated from nearest-seed assignment in the experiments.

\begin{figure*}[t]
\centering
\includegraphics[width=.82\textwidth]{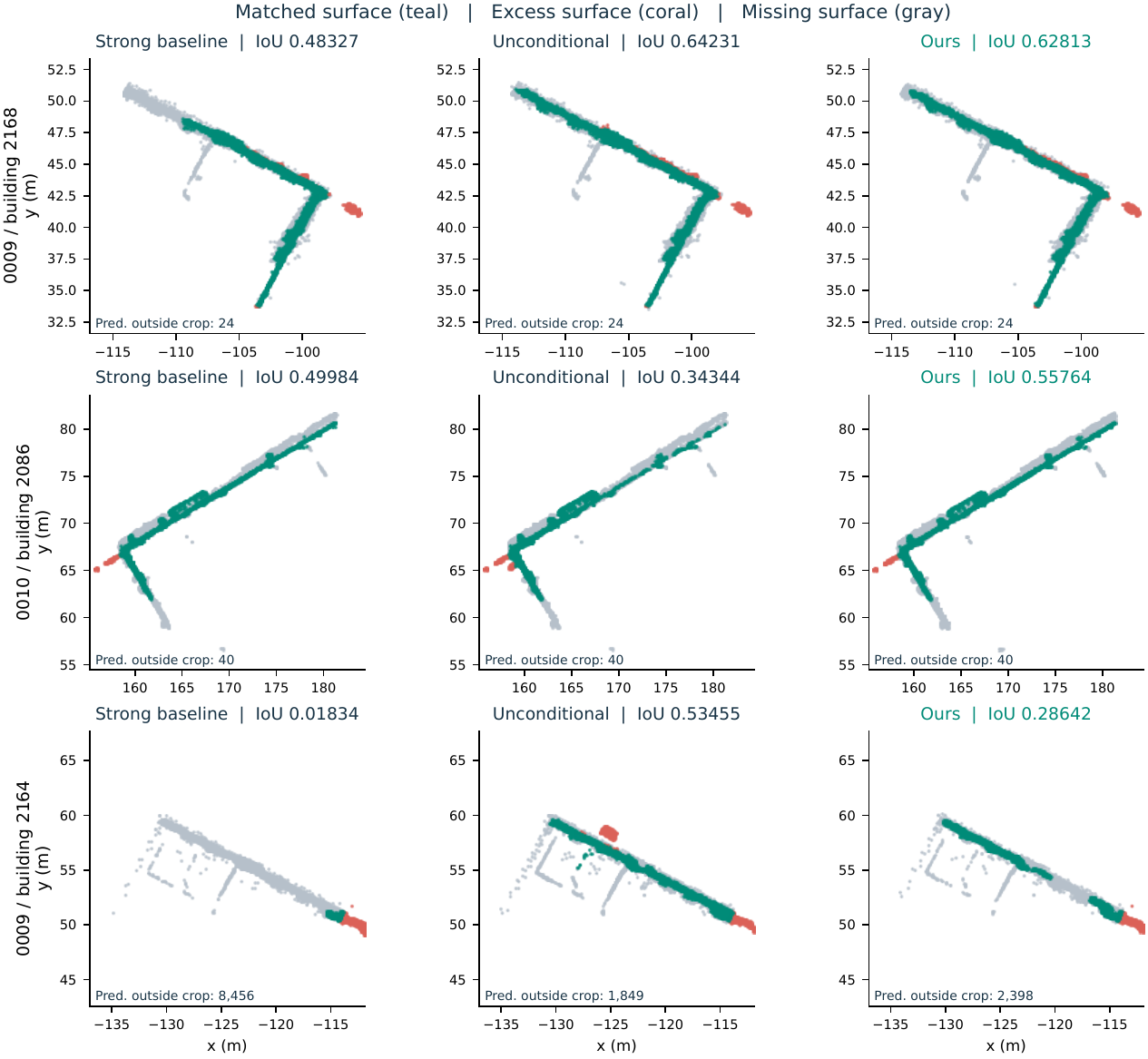}
\caption{Two improved building reconstructions and a rejected useful update. Teal denotes correct target support, coral excess surface, and gray missed surface. Crops share bounds within each row; IoU is computed over the full confidence-weighted reference domain. For building 2164, unconditional revision reaches IoU 0.535 and Ours reaches 0.286. These cases were selected after scoring to examine the revision decisions.}
\label{fig:cases}
\end{figure*}

\section{Experiments}
\subsection{Data and evaluation}
We evaluate building and car instances on KITTI-360~\cite{kitti}. The prospective comparison uses sequence 0009, frames 778--1026 (50 sampled image--LiDAR pairs), and sequence 0010, frames 1872--2033 (34 pairs). Seven development-exposed sequences were excluded in full, and the windows and method were fixed before image processing and semantic-reference decoding. A second comparison uses four spatial holdouts from previously exposed sequences 0003--0006, with 61 pairs each. These two groups are reported separately. Geometry weights were fitted on development data from 0000/0002 and remain fixed throughout.

We transfer predicted IDs to the full static reference domain using the nearest map point within a strict 0.20\,m radius. A local confidence-weighted adapter calls the public AP kernel with the upstream train-ID reference encoding. AP is averaged over IoU thresholds 0.50, 0.55, $\ldots$, 0.90, with a minimum reference confidence mass of 100. The two prospective windows contain 83 eligible instances, including uncovered objects. All AP values are percentages. This protocol evaluates local maps and is not an official test-server submission. Appendix~\ref{app:protocol} gives the selection and scoring details.

\paragraph{Controls.}
The strong baseline is the frozen object map. Unconditional revision applies all shared component assignments. History-only applies the authorized old-label changes while keeping every parent-added point, label, and score. \ours{} also applies the local admission and assignment policy. These controls share perception outputs and surface proposals. A MaskClustering core adaptation is reported separately because its mask inputs and semantic readout differ from the native method.

\subsection{Map accuracy}
On the prospective windows, \ours{} reaches 18.762 mean AP, compared with 17.259 for the strong baseline and 17.712 for history-only (Table~\ref{tab:results}a). AP50 rises from 33.453 to 34.319 in 0009 and from 18.056 to 30.556 in 0010. The canonical-class and uniform-score checks also improve, indicating that the gain is not solely a change in ranking scores (Appendix~\ref{app:protocol}).

Two additional buildings, 2168 and 2086, cross the AP50 threshold without losing an existing match. Both were partially covered by the baseline. For 2086, IoU moves from 0.49984 to 0.55764, so one added match starts immediately below the threshold. Figure~\ref{fig:cases} shows these gains alongside building 2164, for which unconditional revision produces a better completion.

On the four spatial holdouts, mean AP increases from 20.263 to 23.171. Four canonical AP50 matches are gained and none are lost. However, unconditional revision reaches 23.251, and \ours{} falls by 0.557 AP points relative to the baseline in 0004. The gains over the backbone therefore extend to another set of locations, while the relative benefit of conservative revision varies across windows.

\subsection{Effects of revision decisions}
We use five post-hoc output interventions per prospective window to separate historical relabeling from new-point assignment. Each copies fixed predictions and changes only the designated labels or admitted sources. Table~\ref{tab:causal} compares paired interventions; the complete definitions and scores are in Appendix~\ref{app:causal}.

\begin{table}[t]
\centering\footnotesize
\begin{tabular}{@{}lrrr@{}}
\toprule
AP difference (pp) & 0009 & 0010 & Mean \\
\midrule
Apply authorized old edits & +0.161 & +0.000 & +0.081 \\
Protect vs. unconditional & +0.073 & +2.778 & +1.426 \\
Local policy vs. witness & +0.097 & +1.389 & +0.743 \\
New certificate vs. nearest & +0.000 & +0.000 & +0.000 \\
\bottomrule
\end{tabular}

\caption{AP differences between paired output interventions. Historical-label contrasts hold new coordinates, labels, and scores fixed; new-point contrasts hold historical decisions fixed. Rows are conditional effects and are not additive.}
\label{tab:causal}
\end{table}

\paragraph{Historical edits and preservation.}
In 0009, applying the authorized old-label changes adds 0.161 AP points relative to preserving the old labels at the same new geometry. A fixed baseline-to-reference correspondence identifies the changes as consolidation of fragments belonging to the same building. In 0010, \ours{} changes no historical labels, yet preserving them instead of applying unconditional edits adds 2.778 points. Protecting an existing assignment can thus account for a gain even when no historical edit is executed.

\paragraph{New-point assignment.}
Local assignment improves mean AP by 0.743 points over direct mask-supported admission with parent labels. Removing the extra seed-frame stability requirement gives exactly the same AP as \ours{} in all six evaluation windows. In 0009, the check excludes 35 additional sources and changes five reference labels; in 0010, it changes neither. The observed benefit comes from local identity assignment, with no measured AP increment from the additional check.

\subsection{Failure cases and scope}
Conservative merging leaves some correct completions unfinished, as building 2164 illustrates. On 0009, best same-class object IoU improves for 15 canonical objects, declines for 26, and is unchanged for 21; the corresponding counts in 0010 are 2, 0, and 10. Unchanged AP50 match counts therefore conceal some geometric regressions. Car AP in 0010 remains 6.173 for all four primary methods, and car AP in 0009 falls from 45.411 to 45.355. Building improvements dominate the aggregate gain.

Development results are also reported in Appendix~\ref{app:development}: mean AP is 28.012 for the strong baseline, 28.048 for \ours{}, and 28.217 for history-only. That comparison did not meet the original development criterion. The prospective evaluation and spatial extension retain their separate selection rules and outcomes.

\subsection{Garage integration}
We applied the same revision code to 330 frames of \texttt{rosbag2245}, using the existing LIVO poses and point archive. \ours{} produces 717,816 points and 68 nonempty IDs while retaining all old labels. Appendix~\ref{app:robot} shows the four-method comparison with recorded images. This offline run uses the complete measurement archive; the public-data adapter respects source acquisition time. Without independent garage annotations, the comparison demonstrates integration but does not measure semantic accuracy.

\section{Conclusion}
Historical relabeling and surface expansion have different failure modes. Testing them separately improves the evaluated object maps through selective fragment consolidation, identity preservation, and local assignment of new measurements. The policy can also reject useful merges, and its additional seed-frame check provides no AP gain over nearest-seed assignment. Evaluation remains limited to local KITTI-360 windows and an unannotated robot recording. Native full-system comparisons, longer sequences, and pose/depth perturbations are needed to assess broader applicability. The current proposal representation also limits recovery: an incorrect old footprint can prevent creation of a new identity inside that component.

\balance
\begingroup
\small
\setlength{\bibsep}{1pt}
\bibliographystyle{unsrtnat}
\bibliography{references_current}
\endgroup

\clearpage\nobalance\onecolumn
\appendix
\begin{center}\Large\bfseries Supplementary Material\end{center}

\section{Surface Proposals and Assignment Details}\label{app:proposals}
For component $C$ and selected query set $\mathcal Q$ in frame $t$, let $U_{t,\mathcal Q}$ be the union of query masks over the visible measured-source domain $V_t$. A visible cell contributes unit mass, divided among its source pixels through weights $w_{tp}$. The shared proposal score is
\begin{align}
F_t(C,\mathcal Q)={}&\sum_{p\in V_t}w_{tp}\,[p\in U_{t,\mathcal Q}]
             (2[p\in C]-1) \nonumber\\
 &-\sum_{q\in\mathcal Q}\left[\log_2(1+Q_t)-\log_2 s_{tq}\right],
\end{align}
where $Q_t$ is the number of positive-score native queries of the same class. Queries are added greedily while their marginal gain is positive; overlapping masks count a source once. A component is accepted when its pooled utility exceeds $\log_2(K+1)$, where $K$ is the number of geometric components, and it has positive utility in at least three frames. All internal controls use the same accepted components and receiver proposals.

The graph uses 0.20\,m cells, 16-neighbor proposals, and a strict 0.40\,m edge bound. Old-ID footprints require a pure cell. Mixed cells supply no historical identity evidence. A component with old footprints but no usable seed abstains. Mask support is evaluated on measured projections; unobserved space is excluded from the evidence domain. Source additions within 0.05\,m of prior geometry are rejected by the shared proposal stage.

\paragraph{Stability under seed-frame removal.}
Equation~\ref{eq:certificate} requires two distinct frame groups whose winning-identity seeds are closer than every competing identity. Removing one group leaves at least one winning seed. Removing competing seeds can only increase their nearest distance. The winning label is therefore unchanged under deletion of one seed-frame group, provided the graph and edge weights remain fixed. Pose error, graph changes, and incorrect masks are outside this guarantee.

\section{Evaluation Protocol}\label{app:protocol}
\paragraph{Prospective windows.}
Sequences 0009 [778,1026] and 0010 [1872,2033] contain 50 and 34 sampled image--LiDAR pairs. Metadata selection excluded all seven sequences used during development. The full inputs contain 6,007,585 measured sources, with a minimum raw-measurement bound separation of 1,140.14\,m from prior inputs and between the selected windows. Method settings and windows were fixed before processing their images; the eight primary maps were saved before semantic-reference decoding.

The primary adapter follows the upstream train-ID reference encoding and confidence weighting. Nearest-source transfer uses a strict 0.20\,m radius. A canonical raw-class check voids noncanonical aliases while retaining positions and weights. A uniform-score check replaces prediction scores with one. The primary domain has 83 eligible instances and the canonical domain has 74, including uncovered instances in both cases.

\begin{minipage}[t]{.48\textwidth}\vspace{0pt}\centering\small
\begin{tabular}{lrrr}
\toprule
Method & Primary & Canonical & Uniform \\
\midrule
Strong baseline & 17.259 & 18.716 & 17.079 \\
Unconditional & 17.050 & 18.546 & 16.486 \\
History-only & 17.712 & 19.428 & 17.111 \\
\textbf{Ours} & 18.762 & 20.727 & 18.905 \\
\bottomrule
\end{tabular}

\captionof{table}{Prospective-window means under the primary and two diagnostic protocols. AP50:90 in percent.}
\end{minipage}\hfill
\begin{minipage}[t]{.48\textwidth}\vspace{0pt}\centering\small
\begin{tabular}{lrrrr}
\toprule
Method & 09 Bldg. & 09 Car & 10 Bldg. & 10 Car \\
\midrule
Strong baseline & 0.785 & 45.411 & 16.667 & 6.173 \\
Unconditional & 1.234 & 44.126 & 16.667 & 6.173 \\
History-only & 1.105 & 44.126 & 19.444 & 6.173 \\
\textbf{Ours} & 1.299 & 45.355 & 22.222 & 6.173 \\
\bottomrule
\end{tabular}

\captionof{table}{Prospective-window per-class AP. Building gains dominate; car AP is unchanged in 0010 and slightly lower in 0009.}
\end{minipage}

\paragraph{Spatial holdouts.}\label{app:extension}
The extension uses 61 uniformly sampled image--LiDAR pairs per window: 0003 [617,738], 0004 [5930,6119], 0005 [2,357], and 0006 [3251,3634]. These windows come from the annotated static train/validation pool, excluding geometry-training sequences 0000/0002. Their selected trajectory bounds are respectively 448.96, 428.48, 542.49, and 359.98\,m from exposed source bounds and earlier selected windows, exceeding the 300\,m metadata threshold. They are spatial holdouts within used sequences. The pool was expanded after a validation-only metadata search found no qualifying windows.

The budget was fixed at four windows, 244 pairs, seven methods, and three scoring protocols. All 28 maps and 84 scores were completed without threshold tuning or window replacement. Window 0003 has one eligible building and no eligible car; its mean uses the building class alone. The four-window mean weights windows equally. Ours performs 44,276 historical source-label edits, gains four canonical AP50 matches, and loses none; all seven methods appear in Table~\ref{tab:results}b.

\section{Development and Selection}\label{app:development}
The exposed development windows in sequences 0000/0002 yielded the results below. Identity-only$^*$ also changes local labels of new points. History-only changes only the historical-label prefix and is therefore the cleaner control for that decision. ``History + witness'' keeps parent labels on mask-supported additions.

\begin{center}\small
\begin{tabular}{lrrr}
\toprule
Method & 0000 & 0002 & Mean \\
\midrule
Strong baseline & 23.674 & 32.349 & 28.012 \\
Unconditional & 20.247 & 32.326 & 26.287 \\
Identity-only$^*$ & 24.193 & 32.040 & 28.116 \\
Write-only & 19.840 & 32.647 & 26.243 \\
No history edits & 23.710 & 32.154 & 27.932 \\
\textbf{Ours} & 23.943 & 32.154 & 28.048 \\
History-only & 24.193 & 32.241 & 28.217 \\
History + witness & 23.741 & 32.355 & 28.048 \\
\bottomrule
\end{tabular}

\captionof{table}{Development AP50:90. All evaluated variants are retained. These results are separate from the prospective windows and spatial holdouts.}
\end{center}

The development criterion was not met: Ours improves by only 0.036 mean AP points over the baseline and remains below history-only. The prospective comparison was registered as a bounded replication after that result. Its acceptance rule required finite scores in both windows, per-window AP/AP50 nonregression, at least one mean AP point over the baseline, a higher mean than both controls, an added canonical AP50 match without lost matches, and nonzero historical edits. The two prospective windows satisfy the rule. The spatial extension fails the per-window nonregression and control-superiority conditions. No window was substituted and no per-window winner was chosen as Ours.

The multi-mask cover and best-single-mask control accept the same number of components in the prospective windows (46 and 5). Higher coverage utility on individual frames therefore does not establish an independent map-level gain. During development, history + witness and Ours also have nearly equal means (28.047974 and 28.048252).

\section{Controlled Output Interventions}\label{app:causal}
The five interventions per prospective window were specified after the primary results were inspected and before intervention scoring. They hold weights, thresholds, proposals, and original predictions fixed. Their effects are conditional on those decisions.

\paragraph{Historical-label controls.}
``Ours new'' copies Ours appended sources, coordinates, labels, and ranking metadata. Replacing only the historical-label prefix with either baseline or unconditional labels measures the effect of old assignments at fixed new geometry. Old-point coordinates and ordering are unchanged.

\paragraph{New-point controls.}
``Witness'' keeps directly mask-supported parent additions with their parent labels, paired with either preserved or authorized history. ``Nearest'' fixes Ours history and assigns supported additions to the unique nearest seed identity without the second seed-frame requirement on new points. Historical authorization, including its local check, remains fixed. All controls share the Ours classification metadata and postprocessing.

\begin{center}\small
\begin{tabular}{@{}lrrr@{}}
\toprule
Export intervention & 0009 & 0010 & Mean \\
\midrule
Unconditional old IDs + Ours new & 23.253 & 11.420 & 17.337 \\
Preserved old IDs + Ours new & 23.166 & 14.198 & 18.682 \\
Preserved old IDs + witness & 23.089 & 12.809 & 17.949 \\
Authorized old IDs + witness & 23.230 & 12.809 & 18.019 \\
Authorized old IDs + nearest & 23.327 & 14.198 & 18.762 \\
\textbf{Ours (fixed)} & 23.327 & 14.198 & 18.762 \\
\bottomrule
\end{tabular}

\captionof{table}{All controlled output interventions, AP50:90 in percent. Only the operations specified above vary.}
\end{center}

For 0009, the nearest-seed control retains 35 sources rejected by Ours and differs at five reference positions, yet AP is equal. For 0010, its source set, labels, and reference predictions are identical to Ours. The witness-history versus witness-preservation contrast is +0.141 / +0.000 AP points in 0009/0010. These paired effects depend on which additions are held fixed and need not sum to the total baseline-to-Ours change.

\subsection{Interpreting historical edits}
Each baseline predicted ID is matched to the same-class canonical reference instance with maximum confidence-weighted intersection; ties use the lower reference ID. Multiple fragments may match the same instance, and IDs without same-class overlap remain unmapped. This correspondence is then held fixed across variants. Each reference point retains its nearest baseline source within 0.20\,m, and only the historical label at that source changes.

For mapped endpoints, wrong-to-correct transitions are corrections, correct-to-wrong transitions are damage, and correct-to-correct transitions consolidate fragments of the same reference object. Wrong-to-wrong transitions remain unresolved. Missing endpoint correspondences and void reference points are reported separately. Counts other than ``Old edits'' are confidence-weighted reference mass, rather than source-point accuracy.

\begin{center}\footnotesize
\begin{tabular}{@{}llrrrrrr@{}}
\toprule
Window & Method & Old edits & Correction & Damage & Consolidation & Unresolved & Unknown \\
\midrule
0009 & Unconditional & 26,788 & 101.1 & 1072.9 & 10848.0 & 7152.8 & 3348.2 \\
0009 & \textbf{Ours} & 12,855 & 0.0 & 0.0 & 10671.7 & 2.7 & 0.0 \\
0010 & Unconditional & 3,427 & 0.0 & 0.0 & 0.0 & 0.0 & 3552.2 \\
0010 & \textbf{Ours} & 0 & 0.0 & 0.0 & 0.0 & 0.0 & 0.0 \\
\bottomrule
\end{tabular}

\captionof{table}{Historical outcomes under a fixed baseline-to-reference correspondence. Unknown endpoints are excluded from correct/incorrect categories.}
\end{center}

The authorized 0009 transitions are $27\!\to\!17$ (8,745 sources, building 2168) and $36\!\to\!48$ (4,110 sources, building 2486). Both consolidate fragments of the same object; 2.736 reference mass remains wrongly assigned. The unconditional 0010 transition targets ID 6, which is absent from the baseline correspondence, leaving 3,552.226 reference mass unknown in this diagnostic. Its effect on AP is measured by the fixed-new-geometry intervention in Table~\ref{tab:causal}.

\section{Runtime and External Comparison}\label{app:runtime}
\subsection{Revision runtime}
On an AMD Ryzen 9 9950X, cached graph assignment takes 117.66\,ms for 0009 and 17.33\,ms for 0010. The graph timer includes both native multi-source distance searches and the certificate calculation. The pair timer evaluates merge decisions from cached per-view evidence. Model inference, projection, proposal construction, loading, and export are excluded. These are single CPU runs rather than end-to-end latency estimates.

\begin{center}\small
\begin{tabular}{@{}lrrrrrr@{}}
\toprule
Window & Nodes & Edges & Seeds & Graph (ms) & Pairs (ms) & Prior prep. (s) \\
\midrule
0009 & 42,955 & 457,332 & 61,886 & 117.66 & 0.180 & 5.201 \\
0010 & 15,091 & 129,426 & 11,665 & 17.33 & 0.001 & 2.323 \\
\bottomrule
\end{tabular}

\captionof{table}{Revision input sizes and component timings. Edges are directed. Prior preparation was measured in a separate run and is not added to the replay timers.}
\end{center}

The replay driver's peak resident memory is 198.4\,MiB across the two cached runs. Map export takes 0.286--0.290\,s for 0009 and 0.088--0.089\,s for 0010. Native distances and certificate arrays reproduce the archived results exactly. The geometry weights were fitted on exposed 0000/0002 data; neither the replay nor the extension refits them.

\subsection{MaskClustering core adaptation}\label{app:external}
We run the official MaskClustering graph construction, iterative view-consensus clustering, and postprocessing code through a CPU adapter. Six upstream files are pinned; only device substitutions and an unused backprojection import are changed. Inputs are disjoint highest-score SAM3 query supports on the same measured sources, using the causal visibility cache. Query scores provide the building/car readout. This replaces the native CropFormer~\cite{cropformer}/OpenCLIP~\cite{openclip} frontend and also differs from the internal backbone's semantic readout.

We retain the upstream ScanNet defaults: visible and undersegmentation ratios 0.3, consensus 0.9, contained-mask ratio 0.8, point-filter ratio 0.5, DBSCAN radius 0.1\,m with four minimum points, two supporting masks, and overlap merge ratio 0.8. Empty graphs export no clusters. Source ties use stable object order in the nearest-map readout.

\begin{center}\small
\begin{tabular}{@{}lrrr@{}}
\toprule
Window & Primary AP & Canonical AP & Uniform-score AP \\
\midrule
0009 & 0.0071 & 0.0082 & 0.1664 \\
0010 & 0.0000 & 0.0000 & 0.0000 \\
\bottomrule
\end{tabular}

\captionof{table}{MaskClustering core adaptation on the prospective windows. AP is reported in percent; the four-window extension is included in Table~\ref{tab:results}b.}
\end{center}

The adapter yields 2,542 / 662 intermediate clusters and 98 / 23 final objects on 0009/0010. Fragmentation is severe and AP is low. Because the inputs and readout have changed and indoor defaults are retained, this comparison diagnoses the adaptation; it cannot establish superiority over the native MaskClustering pipeline.

\clearpage
\section{Image Projections and Complete Maps}\label{app:maps}
The figure compares the same four primary outputs over each complete prospective window. All 74 canonical and 83 primary reference objects remain in the accompanying per-object result file. For the qualitative cases in Figure~\ref{fig:cases}, full-domain IoUs are listed below.
\begin{center}\small\begin{tabular}{lrrrr}
\toprule
Window / object & Strong & Uncond. & History & Ours \\
\midrule
0009 / 2168 & 0.48327 & 0.64231 & 0.64354 & 0.62813 \\
0010 / 2086 & 0.49984 & 0.34344 & 0.50627 & 0.55764 \\
0009 / 2164 & 0.01834 & 0.53455 & 0.38001 & 0.28642 \\
\bottomrule
\end{tabular}
\end{center}

\begin{center}
\includegraphics[width=.95\textwidth]{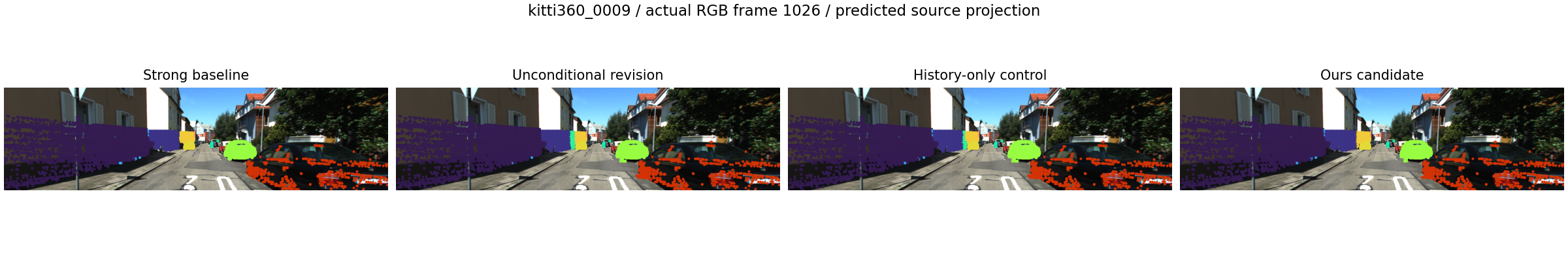}
\includegraphics[width=.95\textwidth]{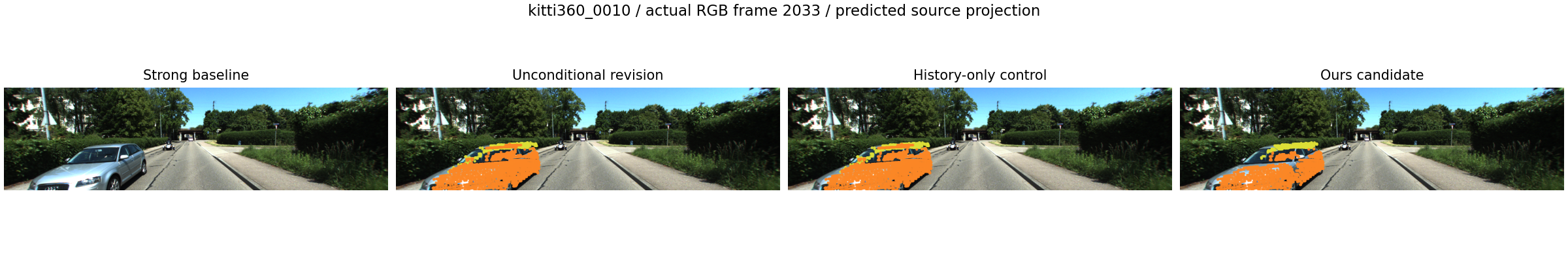}
\includegraphics[width=.90\textwidth]{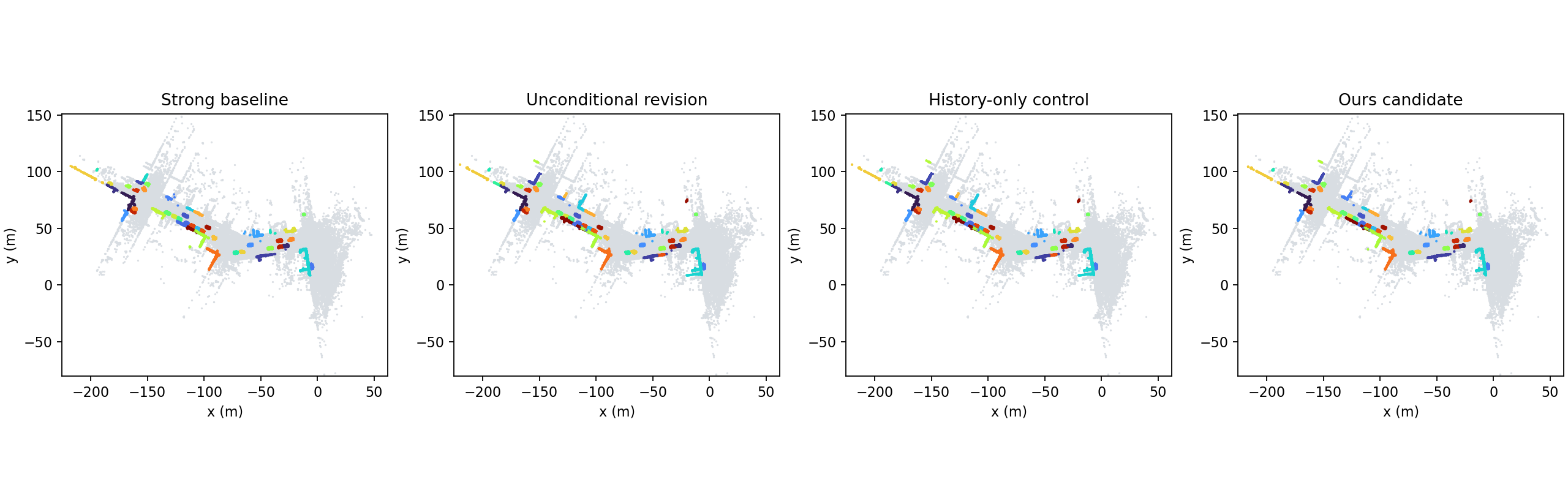}
\includegraphics[width=.90\textwidth]{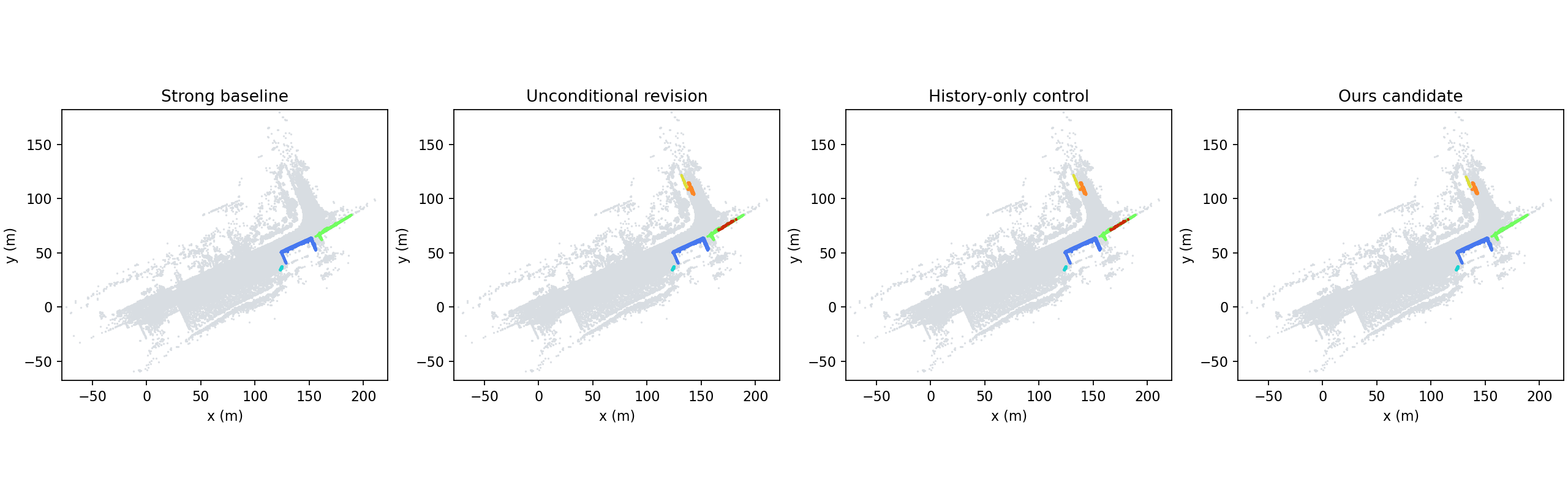}
\captionof{figure}{Prospective KITTI-360 windows. Top: recorded final sampled images and visible measured-source projections for baseline, unconditional, history-only, and Ours. Bottom: the corresponding full-window object maps. Colors denote predicted IDs; gray denotes measured context. Image display uses a 0.05\,m transfer radius, while evaluation uses 0.20\,m. Map display uses fixed-stride subsampling of at most 60,000 context and 40,000 object points; scores use all exported points. Bounds are shared within each window.}
\label{fig:maps}
\end{center}

\clearpage
\section{Garage Integration}\label{app:robot}
All four methods use the same 330 recorded frames and LIVO outputs. Ours authorizes no historical donor merge in this run. The table counts all exported points and nonempty IDs; the figure's map panels show the vehicle-class subset. The recording has no independent contours or cross-frame object annotations, so these outputs are qualitative integration evidence.
\begin{center}\small\begin{tabular}{lrrr}
\toprule
Method & Points & IDs & Old edits \\
\midrule
Strong baseline & 65,189 & 61 & 0 \\
Unconditional & 1,158,094 & 69 & 11,100 \\
History-only & 1,158,094 & 69 & 0 \\
\textbf{Ours} & 717,816 & 68 & 0 \\
\bottomrule
\end{tabular}
\end{center}
\begin{center}
\includegraphics[width=.91\textwidth]{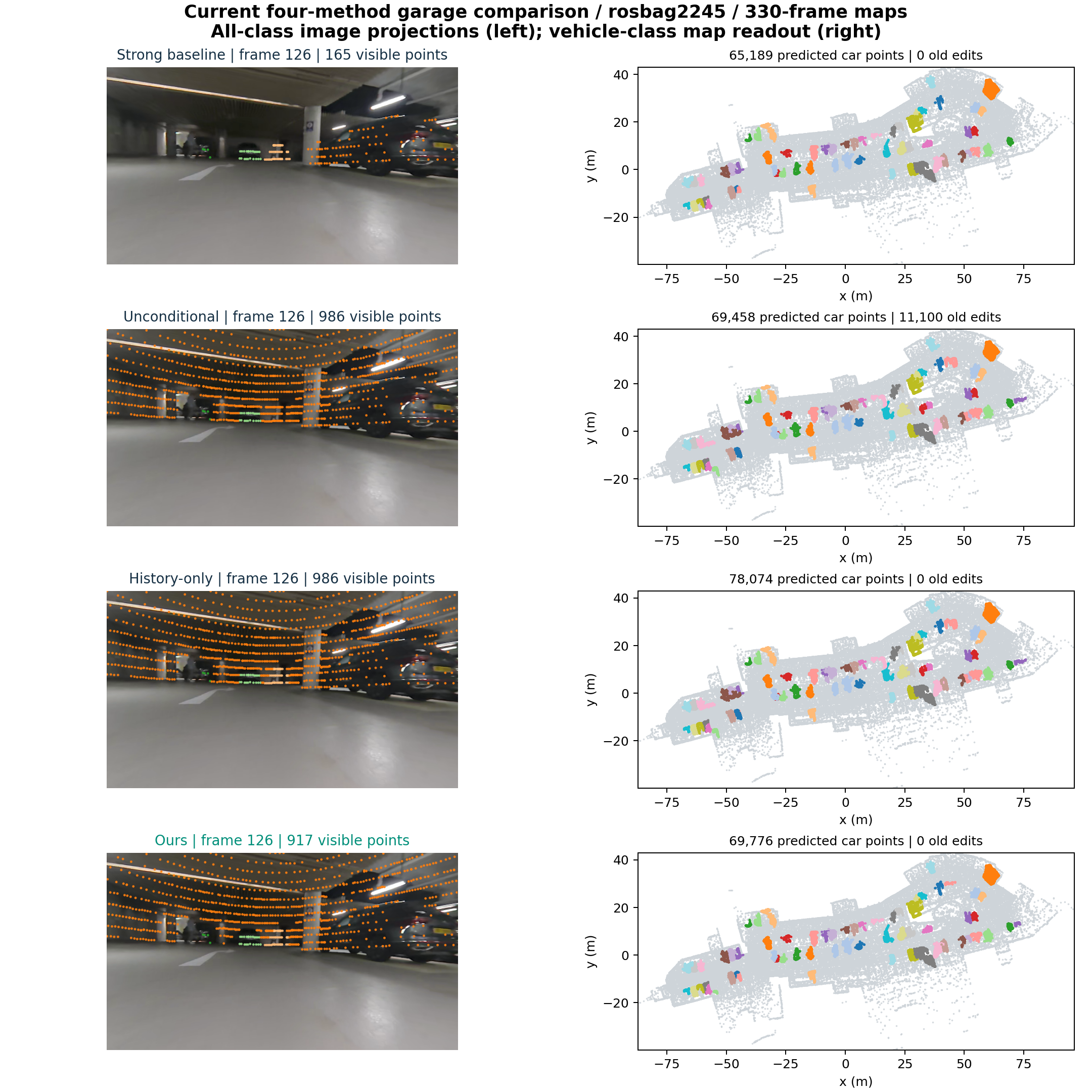}
\captionof{figure}{Garage frame 126 and the corresponding object maps. Left: RGB with measured-depth-gated point projections. Right: predicted vehicle points within shared map bounds. Context is subsampled to 60,000 points; vehicle points are unthinned. Vehicle counts are 65,189 / 69,458 / 78,074 / 69,776 for baseline / unconditional / history-only / Ours. Colors distinguish predicted IDs within each map and do not establish cross-method physical correspondence. The offline run uses the full measurement archive.}
\label{fig:robot-current}
\end{center}

\paragraph{Reproducibility and tool use.}
Tables and visualizations use saved predictions; protocol hashes and per-object results accompany the source. OpenAI Codex assisted with implementation, experiment orchestration, result inspection, literature checks, figures, and writing. Photographs and point clouds are recorded data.

\end{document}